\documentclass[10pt,twocolumn,letterpaper]{article}

\usepackage[pagenumbers]{cvpr} 

\usepackage{graphicx}
\usepackage{amsmath}
\usepackage{amssymb}
\usepackage{booktabs}
\usepackage{float}
\usepackage{caption}
\usepackage{multirow}
\usepackage{flushend}
\usepackage{soul}

\usepackage[pagebackref,breaklinks,colorlinks]{hyperref}

\usepackage[capitalize]{cleveref}
\crefname{section}{Sec.}{Secs.}
\Crefname{section}{Section}{Sections}
\Crefname{table}{Table}{Tables}
\crefname{table}{Tab.}{Tabs.}

\begin{document}
	
	\title{Seeing Through the Noisy Dark: Towards Real-world Low-Light Image Enhancement and Denoising}
	
	\author{Jiahuan Ren$^{1,2}$, Zhao Zhang$^{*1,2}$, Richang Hong$^{1,2}$, Mingliang Xu$^3$,Yi Yang$^4$, Shuicheng YAN$^5$	\vspace{0.3cm}\\
		$^1$School of Computer Science and Information Engineering, Hefei University of Technology, China\\
		$^2$Key Laboratory of Knowledge Engineering with Big Data (Ministry of Education) \& Intelligent\\ Interconnected Systems Laboratory of Anhui Province, Hefei University of Technology, China\\
		$^3$School of Information Engineering, Zhengzhou University, China\\
		$^4$School of Computer Science and Technology, Zhejiang University, China\\
		$^5$ Sea AI Lab (SAIL), Singapore
		\vspace{0.3cm}
	}
\maketitle

\begin{abstract}
	\vspace{-0.4cm}
	Low-light image enhancement (LLIE) aims at improving the illumination and visibility of dark images with lighting noise. 
	To handle the real-world low-light images often with heavy and complex noise, some efforts have been made for joint LLIE and denoising, which however only achieve inferior restoration performance.
	We attribute it to two challenges: 1) in real-world low-light images, noise is somewhat covered by low-lighting and the left noise after denoising would be inevitably amplified during enhancement; 
	2) conversion of raw data to sRGB would cause information loss and also more noise, and hence prior LLIE methods trained on raw data are unsuitable for more common sRGB images.
	In this work, we propose a novel Low-light Enhancement \& Denoising Network for real-world low-light images (RLED-Net) in the sRGB color space.
	In RLED-Net, we apply a plug-and-play differentiable Latent Subspace Reconstruction Block (LSRB) to embed the real-world images into low-rank subspaces to suppress the noise and rectify the errors, such that the impact of noise during enhancement can be effectively shrunk.
	We then present an efficient Crossed-channel \& Shift-window Transformer (CST) layer with two branches to calculate the window and channel attentions to resist the degradation (e.g., speckle noise and blur) caused by the noise in input images. 
	Based on the CST layers, we further present a U-structure network CSTNet as backbone for deep feature recovery, and construct a feature refine block to refine the final features.
	Extensive experiments on both real noisy images and public image databases well verify the effectiveness of the proposed RLED-Net for RLLIE and denoising simultaneously. 

\end{abstract}

\vspace{-0.4cm}
\section{Introduction}
\label{sec:intro}

Low-light image enhancement (LLIE)~\cite{Wei2018DeepRD, Zhang2021BeyondBL, Chen2018LearningTS} is an important low-level task to enhance the illumination of low-light images to normal-light ones~\cite{ Wei2018DeepRD,Land1977TheRT,Ren2020LR3MRL, Guo2017LIMELI}, benefiting many visual applications for processing the low-light images with poor visibility, for instance object detection, image recognition and segmentation in the dark~\cite{Li2021LowLightIA}. 
Early methods solve this task by designing minimal reconstruction models~\cite{Land1977TheRT, Pizer1987AdaptiveHE},
which generally have limited capabilities for restoring detailed information of the low-light images.
In recent years, deep neural networks have been widely applied to LLIE tasks~\cite{Liu2021RetinexinspiredUW, Wu2022URetinexNetRD, Zhang2021LearningTC, Lim2021DSLRDS}. 
These deep models are mostly trained on dark images with relatively low noise, achieving good performance on recovering local textures and global structures of the images~\cite{Zhang2022DeepCC, Xu2022SNRAwareLI, Feng2022LearnabilityEF, Li2022LearningTE}. 

Among previous deep LLIE methods, some tackle image enhancement jointly with denoising, such as~\cite{Wei2018DeepRD,Zhang2021BeyondBL, Liu2021RetinexinspiredUW, Wu2022URetinexNetRD} that are aimed at removing noise in some specific layers while enhancing image in other components, or~\cite{Lu2022ProgressiveJL, Zhang2020AttentionBasedNF, Feng2022LearnabilityEF} that are aimed at removing noise and enhance image jointly. Although these methods achieve promising effectiveness, they still tend to give inferior results with speckle noise and blur when applied to real-world low-light sRGB images, as shown in Figure~\ref{fig:1}. 
We hold that such performance gap mainly roots in two challenges of real-world LLIE.
(1) The noise issue in real-world LLIE is rather complex. 
Due to insufficient lighting or hardware limitations, the images captured in real-world dark environments inevitably contain varying degrees of noise, which is often with unknown distribution. Moreover, some noise is somewhat covered up by the low-lighting, which makes it hard to be separated completely with traditional methods. 
Besides, the unobserved noise in the image would be amplified and lead to speckle noise and blur during the process of enhancement.
(2) The existing LLIE methods trained on raw data are unsuitable for sRGB images due to the difference in the domains of raw and sRGB data. The low-light sRGB image contain less information compared with a raw one, and the conversion of raw data to sRGB may also produce more noise.

\begin{figure}[t]
	\centering
	\includegraphics[width=\linewidth]{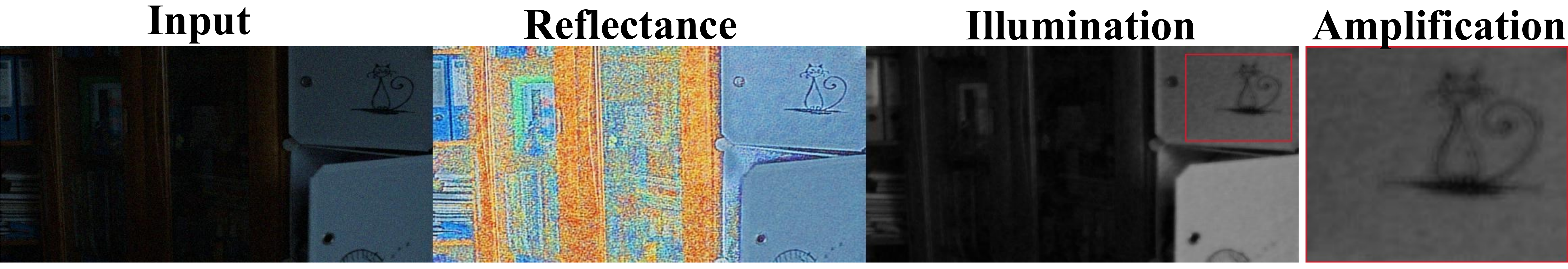}
	\vspace{-0.6cm}
	\caption{Image decomposition via RetinexNet. It can be observed clearly that the illumination of a low-light image contains noise.}
	\label{fig:2}
	\vspace{-0.3cm}
\end{figure}

We conduct a pilot study to investigate the separation of noise from the low-light image~\cite{ Wei2018DeepRD} and provide visualizations in Figure~\ref{fig:2}. 
It can be observed that both reflectance and illumination contain noise, and the final enhanced result (Figure~\ref{fig:1}(b)) contains much speckle noise, which indicates that the noise is hard to be separated completely by the delicately designed layers, and that the hidden noise indeed causes inaccurate representation.

We are then motivated to design a new Low-light Enhancement \& Denoising Network for enhancing real-world low-light images in the sRGB color space, shorted as RLED-Net. 
Instead of removing the noise in some specific layers, we propose to suppress the noise in the low-rank/low-dimensional subspaces which can be used to characterize the real-world images. In RLED-Net, a low-rank subspace recovery block (LSRB) is used to directly embed the noisy dark image into the low-rank subspaces to make our network more suitable for real-world low-light image denoising. 
Furthermore, to reduce the speckle noise and blur caused by the hidden noise in the low-light environment, we propose to borrow some self-attentions to build transformer layers for recovering the important features.
Hence we design a Crossed-channel \& Shift-window Transformer (CST) layer with two branches. Specifically, CST can simultaneously maintain more accurate local features (e.g., edge/texture) and global features (e.g., color/shape), by shift-window attention and crossed channels attention in two parallel branches. Through the ablation studies, We find this structure can preserve more useful features.

We conduct extensive experiments to prove the effectiveness of the proposed RLED-Net. The obtained results demonstrate that our RLED-Net can significantly improve the performance when employed for real-world low-light sRGB images enhancement and denoising. An example can be found in Figure~\ref{fig:1}. 
To sum up, the main contributions of this paper are described as follows:

\begin{itemize}
 \item We propose a plug-and-play and differentiable low-rank subspace recovery block (LSRB), which can suppress the noise in real-world low-light images effectively by representing them in low-rank subspaces.  
 
 \item To reduce the speckle noise and blur caused by hidden noise, we design a two-branch transformer (CST) layer, which can simultaneously recover more accurate local features and global features by self-attention.
 \item We develop a Low-light Enhancement \& Denoising Network (RLED-Net) for enhancing real-world low-light images in the sRGB color space, which is a more widely used color space in reality. SOTA performance of RLED-Net on both well-designed and real low-light images is verified through extensive experiments.
 
\end{itemize}
\begin{figure*}[t]
	\centering
	\includegraphics[width=0.99\linewidth]{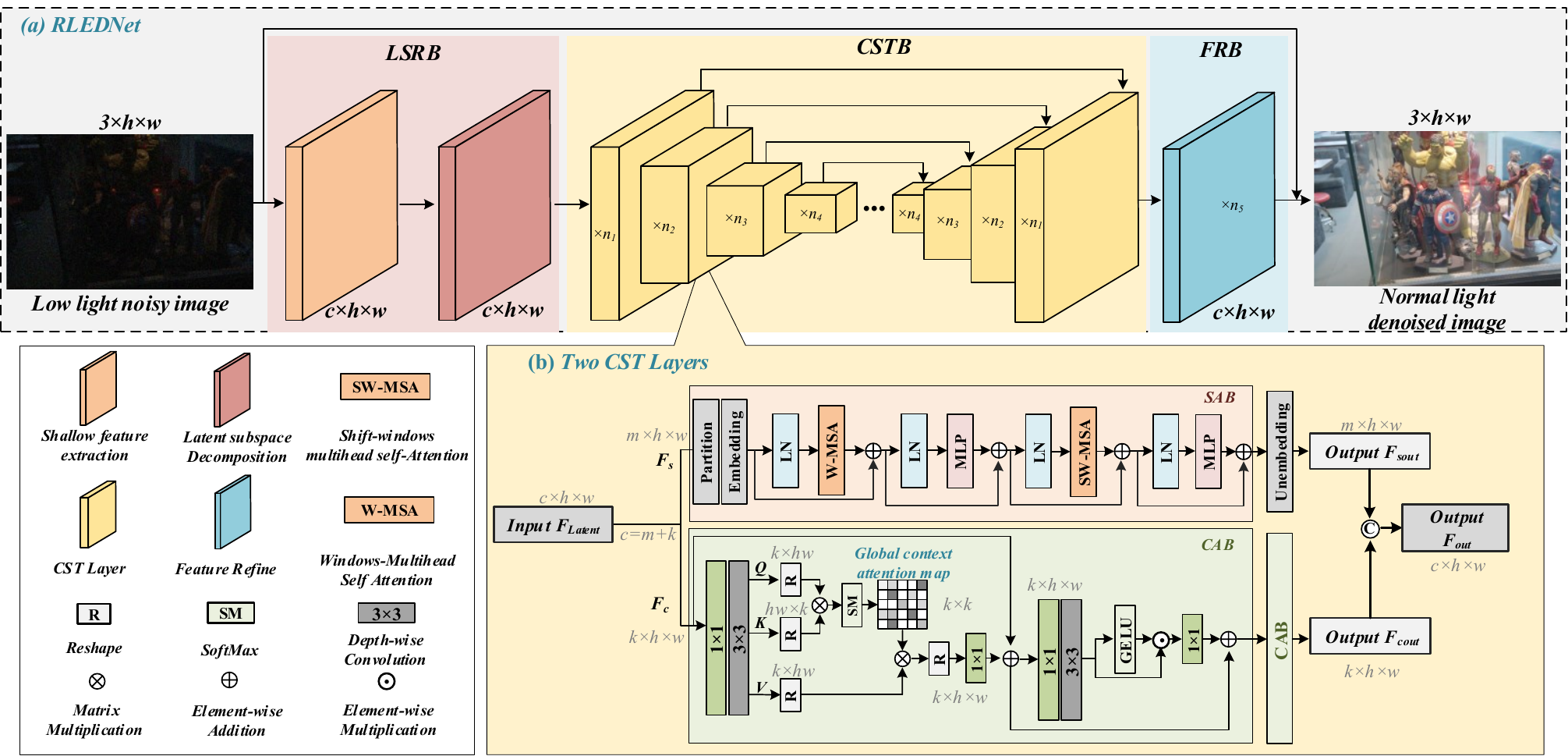}
	\vspace{-0.3cm}
	\caption{The framework of our RLED-Net (a), which consists of three main components, i.e., plug-and-play and differentiable Latent Subspace Reconstruction Block (LSRB), CST based Network (CSTNet) and Feature Refine Block (FRB). Each encoder and decoder of CSTNet include multiple CST layers, and the structure of two CST layers is shown in (b).}
	\label{fig:3}
	\vspace{-0.4cm}
\end{figure*}

\section{Related Work}
\label{sec:formatting}

\subsection{Low-light image enhancement}
Low-light image enhancement (LLIE) explores how to enhance the illumination to improve the visibility of low-light images~\cite{Li2021LowLightIA, Zhang2022DeepCC, Xu2022SNRAwareLI,Jiang2021EnlightenGANDL, Ma2022TowardFF }.
Recently, some LLIE methods have been proposed to exploit the restoration of global and local information for more natural restored results. For example, DCC-Net~\cite{Zhang2022DeepCC} decomposes a color image into a gray image and a color histogram, so as to retain the color consistency between the enhanced image and ground truth; SNR~\cite{Xu2022SNRAwareLI} collectively exploits long-range and short-range operations with Signal-to-Noise Ratio (SNR) prior to dynamically enhance the pixels with spatial-varying operations; LLFlow~\cite{Wang2022LowLightIE} uses a normalizing flow network which takes low-light images/features as the condition and maps the distribution of normal-light images to model the conditional distribution. Due to the difficulty of collecting paired data, some unsupervised LLIE methods have also been developed.
For example, EnlightenGAN~\cite{Jiang2021EnlightenGANDL} applies a dual-discriminator to balance the global and local low-light enhancement, and a self-regularized perceptual loss to constrain the distance between the low-light and enhanced images;  ZeroDCE++~\cite{Li2022LearningTE} formulates the LLIE task as an image-specific curve estimation problem using a deep network; SCI~\cite{Ma2022TowardFF} applies a lightweight cascaded illumination learning process with weight sharing for LLIE tasks.
Generally, these previous LLIE methods focus on the LLIE task itself, without explicitly considering the effects of noise in designing their solutions and hence cannot be well applied to the real-world low-light sRGB images, where the noise is heavy and complex and may seriously interfere with the image enhancement process.

\subsection{Joint LLIE and denoising}
Some RetinexNet-based methods have been proposed to address the noise in low-light images.
Retinex~\cite{Land1977TheRT} assumes that low-light images depend on the reflectance and illumination, where reflectance corresponds to the content in the image and the illumination determines the visibility of the image. Specifically, Retinex decomposes the image into reflectance and illumination as
\begin{equation}
	S=R \circ I,
	\label{eq:1}
\end{equation}
where \emph{S} denotes the observed image, \emph{R} and \emph{I} denote the corresponding reflectance and illumination, and $\circ$ is the element-wise product. To obtain \emph{R} and \emph{I}, RetinexNet-based methods~\cite{ Wei2018DeepRD,Zhang2021BeyondBL, Liu2021RetinexinspiredUW, Wu2022URetinexNetRD} usually train a Decomposition-Net to obtain \emph{R} and \emph{I}, a Reflectance Restoration-Net to refine the reflectance \emph{R} of the low-light image and an Illumination Adjustment-Net to adjust the illumination \emph{I} of the low-light images. 
For noise in the low-light image, it is assumed that the noise only exists in reflectance. Accordingly, some denoising network are designed for reflectance. 
The earliest RetinexNet~\cite{ Wei2018DeepRD} employs BM3D~\cite{Burger2012ImageDC} to remove the noise in reflectance; KinD++~\cite{Zhang2021BeyondBL} uses the illumination to guide the reflectance restoration and only performs denoising on reflectance; R2RNet~\cite{Hai2021R2RNetLI} designs a UNet structure denoising network to suppress the noise in reflectance.

In addition to these RetinexNet-based methods, a few deep end-to-end methods can remove noise to some extent.
A recent work~\cite{Lu2022ProgressiveJL} proposes a network with two branches, which works in different resolution spaces and performs joint enhancement and denoising on raw low light images progressively; \cite{Xu2020LearningTR} addresses enhancing and denoising low-light noisy images in the sRGB color space which applies a noise processing in the low-frequency layer.

\section{Proposed Method}

We elaborate on the proposed low-light enhancement \& denoising network RLED-Net for enhancing real-world low-light images in the sRGB color space in detail. 
An illustration of our RLED-Net framework is given in Figure~\ref{fig:3}.
It includes three main components, i.e., the Latent Subspace Reconstruction Block (LSRB), the CST based Network (CSTNet) and the Feature Refine Block (FRB).

Real-world low-light images often contain not only low-resolution pixels, but also unavoidable noise, which would be inevitably amplified in the LLIE process and result in speckles and blur in the enhanced images.
To address these problems, we use LSRB to represent the dark images with low-rank subspaces to suppress noise, which can effectively reduce its negative impact.
Furthermore, low-resolution pixels and disturbing noise will also lead to loss of details and larger chromatic aberration.
Hence, to obtain more accurate global and local features, we propose a basic feature extraction layer, the CST layer. Based on this CST layer, we design CSTNet and FRB for deep feature extraction and refining.
Next, we introduce these components of the proposed RLED-Net according to the sequence of data flow to make the procedures easier to understand.

\subsection{Overall Pipeline and Loss}
Given a noisy low-light image $X\in\mathbb{R}^{3\times h\times w}$, where \emph{c}, \emph{h} and \emph{w} are the number of channels, height and width respectively, RLED-Net firstly feeds it into LSRB to compute latent features $F_{LSRB}\in\mathbb{R}^{c\times h\times w}$, the process of which can also suppress the noise to some extent.
Then, $F_{LSRB}$ is sent into CSTNet to deliver deep features $F_{CSTNet}\in\mathbb{R}^{2c\times h\times w}$, which includes four encoder-decoder pairs, with each pair consisting of multiple CST layers. 
After that, $F_{CSTNet}$ is passed into FRB, and mapped into the same dimension as the original low-light image to obtain a residual image $F_{FRB}\in\mathbb{R}^{3\times h\times w}$. Finally, the restored image $\widehat{Y}$ can be obtained as $ \widehat{Y} = X + F_{FRB}$. The loss function contains three losses, i.e.,  \emph{ $l_1$ loss} ($l_1$), \emph{ssim loss} ($l_{ssim}$) and \emph{TV loss} ($l_{tv}$), which are expressed as follows: 
\vspace{-0.1cm}
\begin{equation}
	L_{total} = l_1(\widehat{Y},Y) + l_{ssim}(\widehat{Y},Y) + \lambda l_{tv}(\widehat{Y}),
	\label{eq:4}
\end{equation}
where $\widehat{Y}$ is the predicted result and \emph{Y} is the ground-truth. 
\begin{figure}[t]
	\centering
	\includegraphics[width=\linewidth]{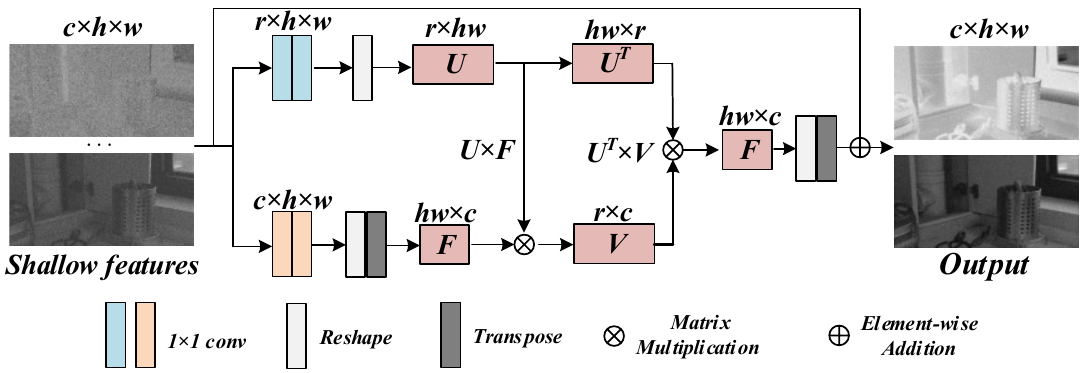}
	\vspace{-0.6cm}
	\caption{The process of latent subspace decomposition.}
	\label{fig:4}
	\vspace{-0.5cm}
\end{figure}
\subsection{Latent Subspace Reconstruction Block (LSRB)}
Images can be represented by low-rank structures~\cite{Ren2022RobustLC, Zhang2020DeepLL, Zhang2021LRNetLS, Ren2021RobustLD}, and the low-rank optimization problem is usually solved by time-consuming and non-differentiable ALM~\cite{Lin2010TheAL} or ADMM~\cite{Boyd2011DistributedOA}.
Inspired by~\cite{Zhang2021LRNetLS}, we design a plug-and-play and differentiable LSRB for latent feature extraction, which effectively reduces the speckle noise or blur in the process of LLIE. 
Specifically, LSRB has a $ 3 \times 3$ convolution for shallow feature extraction and a latent subspace decomposition for feature denoising. As shown in Figure~\ref{fig:4}, several convolutions are used to generate two low-rank matrices: a coefficient matrix \emph{U} and a basis matrix \emph{V}. Specifically, the low-light image $X\in\mathbb{R}^{3\times h\times w}$ is firstly fed into a feature extraction module: 
\vspace{-0.1cm}
\begin{equation}
	F_{shallow} = f_{3\times 3}(X) \in \mathbb{R}^{c\times h\times w},
	\label{eq:6}
\end{equation}
where $f_{3 \times 3}(\centerdot) $ denotes a $ 3 \times 3$ convolution. Then, shallow features $F_{shallow} $ are applied to learn low-rank matrices to approximate $F_{shallow} $ by feeding them to the latent subspace decomposition to obtain the lowest-rank representation. A series of $ 1 \times 1$ convolutions are used for learning the matrices \emph{U} and \emph{V}. As shown in Figure~\ref{fig:4}, the process of learning the coefficient matrix \emph{U} is defined as
\vspace{-0.02cm}
\begin{equation}
	\begin{split}
		& F_U = GELU\left( f_{1 \times 1} \left(  GELU\left( f_{1 \times 1}\left( F_{shallow}\right) \right) \right) \right)\\
		&U = Reshape\left( F_U \right)\in \mathbb{R}^{r\times hw},
	\end{split}
	\label{eq:7}
\end{equation}
where $f_{1 \times 1}(\centerdot) $ is a $ 1 \times 1$ convolution, \emph{GELU($\centerdot $)} is an activation function~\cite{Hendrycks2016GaussianEL}, \emph{Reshape($\centerdot $)} is a flatten operation to convert a 3D matrix to a 2D matrix, and \emph{r} is the rank of matrix. The process of learning the basis matrix \emph{V} is formulated as
\begin{equation}
	\begin{split}
		& F_V = GELU\left( f_{1 \times 1} \left(  GELU\left( f_{1 \times 1}\left( F_{shallow}\right) \right) \right) \right)\\
		& F = Trans \left(Reshape\left( F_U \right)\right)\in \mathbb{R}^{hw \times c} \\
		& V = U \times F \in \mathbb{R}^{r \times c},
	\end{split}
	\label{eq:8}
\end{equation}
where \emph{Trans($\centerdot $)} denotes the transpose operation, and \emph{V} can be regarded as the low-rank representation with the dimension or rank being \emph{r}, where $ r \ll hw$. Finally, the denoised features are obtained as follows: 
\vspace{-0.05cm}
\begin{equation}
	\begin{split}
		& \widehat{F} = U^T \times V \in \mathbb{R}^{hw \times c},\\
		& F_{LSRB} = Trans \left(Reshape^T\left( \widehat{F} \right)\right)\in \mathbb{R}^{c \times h \times w}, 
	\end{split}
	\label{eq:9}
\end{equation}
where $Reshape^T$ is used to convert a 2D matrix back to a 3D matrix. Then, the denoised features $F_{LSRB}$ are transferred into CSTNet, which can also suppress the impact of the noise on the restoration process to a certain extent. 

\subsection{CST based Network (CSTNet)}
The low-resolution pixels of low-light images pose a great challenge to enhancement models, and often fail to provide sufficiently useful information for recovering detailed information and brightness. 
A basic feature extraction layer (CST) with two parallel branches is therefore proposed to tackle this challenge. 
The CST layer has a \emph{Shifted-windows Attention Branch} (SAB) for local feature extraction, and a \emph{Crossed-channel Attention Branch} (CAB) to compute global information. Although SAB can extract local features, it needs to mask some windows to make the self-attention calculation occur between the windows with the same index. Such operations and shifting strategy would result in limited self-attention calculation for the windows located at the image boundary and cause insufficient interaction between windows, leading to loss of important global information~\cite{Liu2021SwinTH, Liang2021SwinIRIR}. Therefore, we also design a convolution based branch CAB. Similar to~\cite{Zamir2022RestormerET}, CAB calculates the self-attention on the crossed channels by the Multi-Head Attention ~\cite{Vaswani2017AttentionIA} to the image restoration task to obtain crossed-channels global context, which is beneficial to modeling the long-distance dependency. The schematic diagram of CST is shown in Figure~\ref{fig:3}(b). We stack multiple CST layers to form four encoders and decoders to construct CSTNet. In this study, the number $n_i$ of CST layers in each encoder/decoder is set to four. 
\begin{figure}[t]
	\centering
	\includegraphics[width=0.95\linewidth]{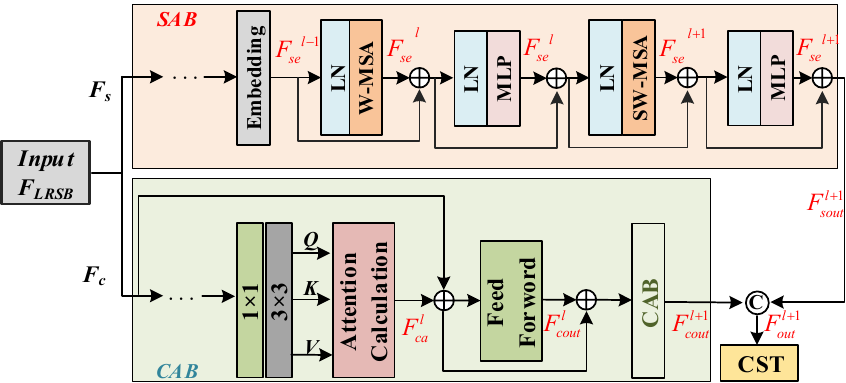}
	\vspace{-0.2cm}
	\caption{The detailed feature calculation process in CST.}
	\label{fig:6}
	\vspace{-0.45cm}
\end{figure}
For the features $F_{LSRB}\in\mathbb{R}^{c\times h\times w}$ , CST divides them into two parts:  $F_{s}\in\mathbb{R}^{m\times h\times w}$ and $F_{c}\in\mathbb{R}^{k\times h\times w}$, where \emph{m+k=c}. SAB firstly partitions $F_s$ into many windows (tokens) and then flattens them by linear embedding, denoted as $F_{se}$. Then, the process of calculating the windows based self-attention is formulated as follows: 
\vspace{-0.05cm}
\begin{equation}
	\begin{split}
		& \widehat{F}^{l}_{se} = W\underline{~~}MSA\left[ LN \left( F^{l-1}_{se}\right)\right]+F^{l-1}_{se}\\
		& F^{l}_{se} = MLP\left[ LN \left( \widehat{F}^{l}_{se}\right)\right]+\widehat{F}^{l}_{se}\\
		& \widehat{F}^{l+1}_{se} = SW\underline{~~}MSA\left[ LN \left( F^{l}_{se}\right)\right]+F^{l}_{se}\\
		& F^{l+1}_{se} = MLP\left[ LN \left( \widehat{F}^{l+1}_{se}\right)\right]+\widehat{F}^{l+1}_{se},\\
	\end{split}
	\label{eq:10}
\end{equation}
where \emph{MLP} and \emph{LN} are the Multilayer Perceptron~\cite{Vaswani2017AttentionIA} and LayerNorm~\cite{Dosovitskiy2021AnII}, respectively. \emph{$W\underline{~~}MSA$} and \emph{$SW\underline{~~}MSA $} are the window based multi-head self-attention and shifted window partitioning configurations, respectively. \emph{l} denotes the \emph{l-th} layer in the windows self-attention. $\widehat{F}^{l}_{se} $ and $F^{l}_{se}$  denote the features of \emph{$W\underline{~~}MSA$} and \emph{MLP} in the \emph{l-th} layer. respectively. $\widehat{F}^{0}_{se} $  denotes the embedding features. Finally, the features of the last layer $\widehat{F}^{n_i}_{se} $  will be transformed into the same dimension as that of the original input features, denoted as $F_{sout}\in\mathbb{R}^{m\times h\times w}$.

For CAB, the features $F_c$ are fed into a $1 \times 1 $ convolution for pixel-wise aggregation of the crossed-channel context, and a $3 \times 3 $ depth-wise convolution for encoding channel-wise spatial context. Based on the enriched context, the terms \emph{Query (Q)}, \emph{Key (K)} and \emph{Value (V)} can be obtained as  
\vspace{-0.05cm}
\begin{equation}
	\begin{split}
		& Q,K,V = f^d_{3 \times 3}(f_{1 \times 1}(F_c)),\\
		& Q,K,V\in\mathbb{R}^{k\times h\times w}.
	\end{split}
	\label{eq:11}
\end{equation}

Then, \emph{Q}, \emph{K} and \emph{V} will be reshaped to $ Q\in\mathbb{R}^{k\times hw}$, $ K\in\mathbb{R}^{hw\times k}$ and $ V\in\mathbb{R}^{k\times hw}$, respectively. \emph{Q} and \emph{K} are employed to generate the crossed-channels global context attention maps $ G\in\mathbb{R}^{k\times k}$ by the dot product. The crossed-channel attention based features $F_{ca}$ in the \emph{l-th} layer can be calculated by the following formula: 
\vspace{-0.05cm}
\begin{equation}
	\begin{split}
		& CA(Q,K,V) = Softmax(Q \centerdot K/\alpha)\centerdot V,\\
		& F^l_{ca} = f^d_{3 \times 3}(CA(Q,K,V))+F_c.
	\end{split}
	\label{eq:12}
\end{equation}

Finally, a feed-forward network~\cite{Zamir2022RestormerET} is applied for the final output of CAB by Eqn.~(\ref{eq:13}) to transform the features $F_{ca}$. That is, it uses the depth-wise convolution for learning the local structures of images for effective restoration:
\vspace{-0.05cm}
\begin{equation}
	\begin{split}
		& F_{fd} = f^d_{3 \times 3}(f_{1 \times 1}(LN(F_{ca}))),\\
		& F^l_{cout} = GELU(F_{fd})\bigodot F_{fd}+F_{ca}.
	\end{split}
	\label{eq:13}
\end{equation}

After both $F^l_{sout}$ and $F^l_{cout}$  are calculated, the output features $F^l_{out}$ of the CST layer can be obtained by concatenating the features  $F^l_{sout}$ and $F^l_{cout}$. The detailed feature calculation process in the CST layer is shown in Figure~\ref{fig:6}.

\subsection{Feature Refine Block (FRB)}
After the reconstructed features $F_{CSTNet}$ are obtained by CSTNet, we feed them into the FRB and calculate the final recovered images by generating a residual image, where FRB is also built by stacking several CST layers and a $ 3 \times 3$ convolution. The process is formulated as
\vspace{-0.05cm}
\begin{equation}
	\begin{split}
		& F_{r} = d_{3 \times 3}\left(\underbrace{CST(\cdots CST(F_{CSTNet}))}_{n_{5}} \right), \\
		& \widehat{Y} = X + F_r		
	\end{split}
	\label{eq:14}
\end{equation}
where $n_5$ is the number of CST layers, $ F_{r} $ is the residual image and $  \widehat{Y}$  denotes the restored image. 
\begin{figure*}[t]
	\centering
	\includegraphics[width=\linewidth]{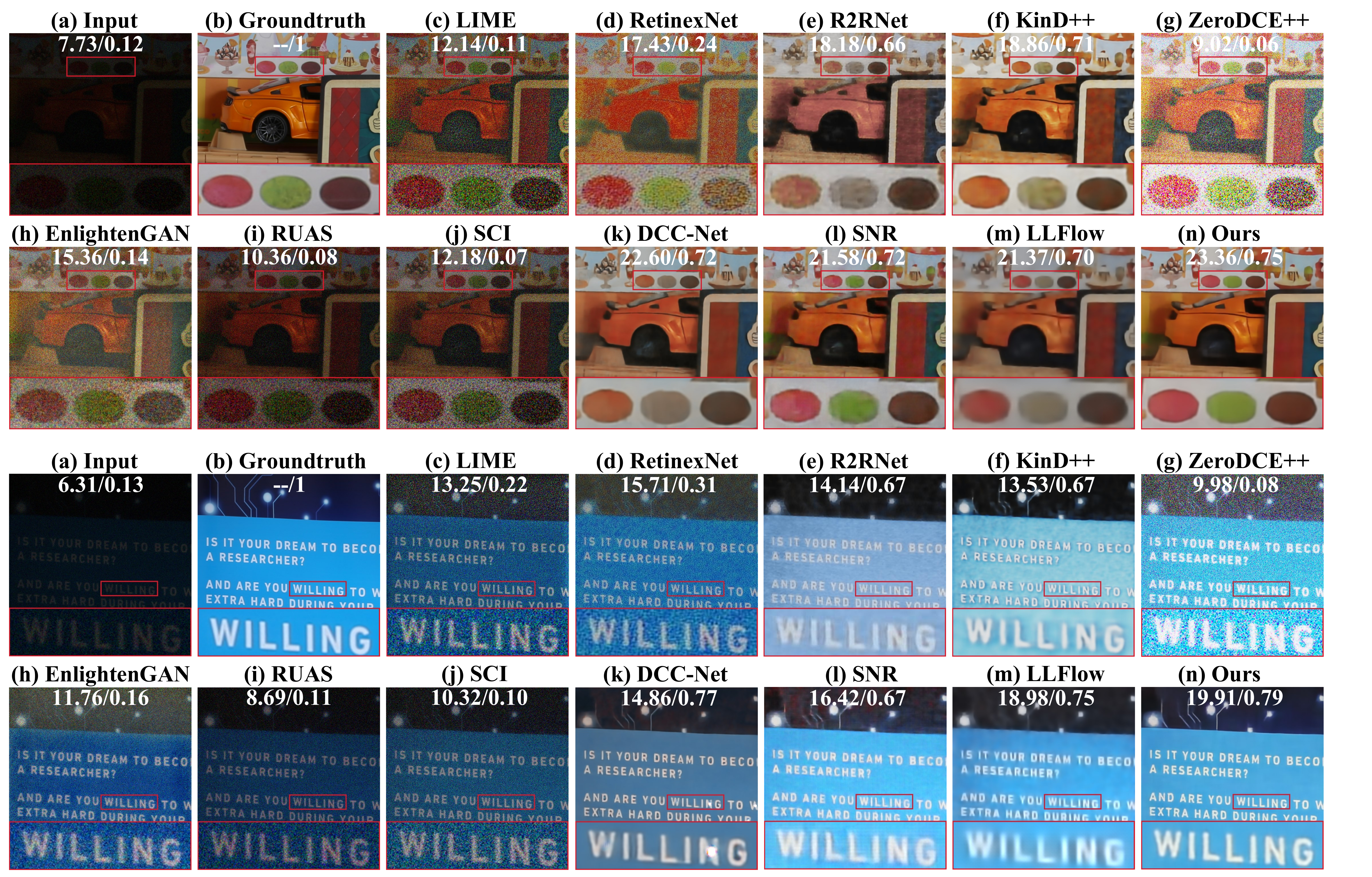}
	\vspace{-0.6cm}
	\caption{Visual comparison of the enhanced images on the RELLISUR dataset (PSNR/SSIM values are also reported).}

	\label{fig:7}
\end{figure*}

\begin{table*}[t]
	\caption{ \centering The numerical results in terms of PSNR, SSIM, CSE ($10^3$) and MAE ($\%$) of all competing methods on the noisy RELLISUR and LSRW datasets ($ \sigma =10$), where the \textbf{bold} denotes the best. }
	\vspace{-0.3cm}
	\centering
	\resizebox{0.99\textwidth}{20mm}{
		\begin{tabular}{ccccccccccccc}
			\hline
			\multirow{2}*{Datasets}  &\multirow{2}*{Eval} & \multicolumn{11}{c}{Methods} \\  \cline{3-13} &~&RetinexNet&R2RNet&KinD++&ZeroDCE++&RUAS&SCI&EnlightenGAN&DCC-Net&SNR&LLFlow&\textbf{Ours}\\
			\hline\hline
			\multirow{4}{*}{RELLISUR} &PSNR$\uparrow$&17.02&19.82&19.05&8.75&10.47&11.69&12.29&21.75&21.73&21.62&\textbf{22.48}\\
			&SSIM$\uparrow$&0.28&0.66&0.69&0.05&0.08&0.08&0.10&0.75&0.72&0.71&\textbf{0.77}\\
			&CSE$\downarrow$&18.08&28.00&27.81&145.21&152.65&148.92&53.67&22.66&8.48&19.07&\textbf{5.71}\\
			&MAE$\downarrow$&11.70&8.61&9.79&30.87&26.25&22.23&20.06&7.03&6.93&6.75&\textbf{6.48}\\
			\hline\hline
			\multirow{4}{*}{LSRW} &PSNR$\uparrow$&16.69&17.24&--&10.03&12.12&13.38&16.09&20.03&20.37&20.21&\textbf{21.63}\\
			&SSIM$\uparrow$&0.25&0.58&--&0.07&0.11&0.08&0.15&0.66&0.64&0.66&\textbf{0.68}\\
			&CSE$\downarrow$&18.28&15.79&--&80.35&80.20&79.70&34.02&7.94&6.49&4.95&\textbf{3.94}\\
			&MAE$\downarrow$&11.99&12.09&--&26.51&21.47&17.99&12.58&7.82&8.38&7.66&\textbf{7.23}\\
			\hline
	\end{tabular}}
	\label{tab1}
\end{table*}
\vspace{-0.1cm}
\begin{figure*}[t]
	\centering
	\includegraphics[width=\linewidth]{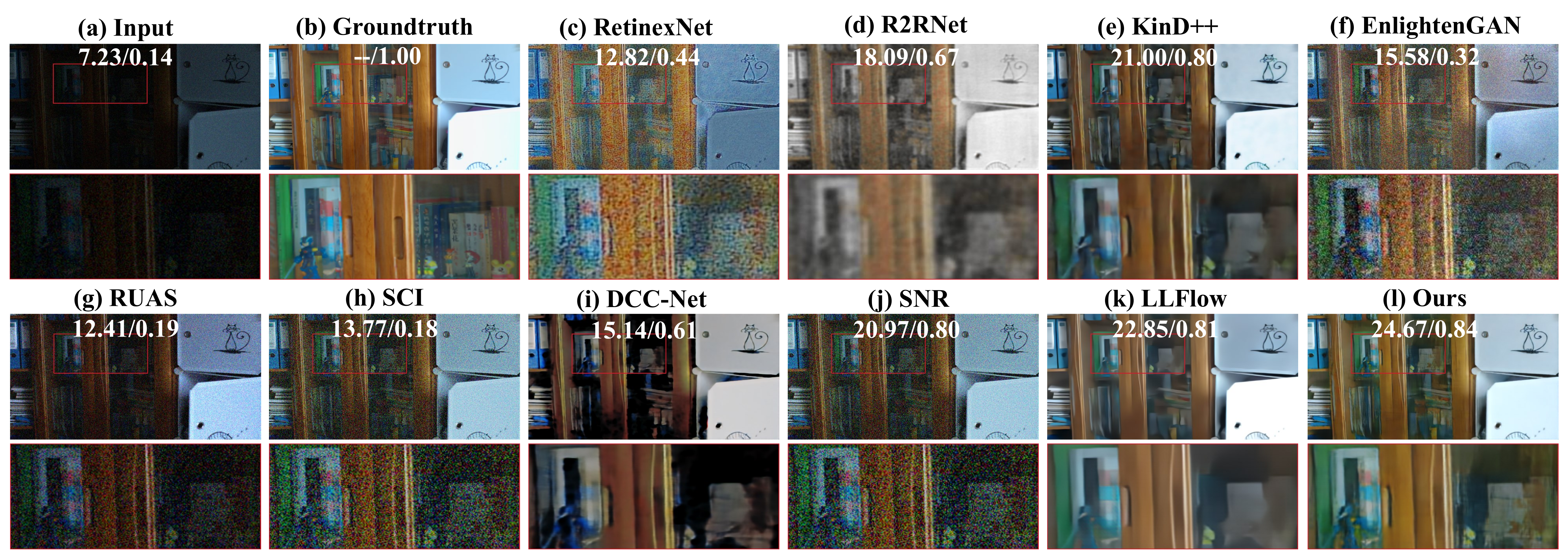}
	\vspace{-0.6cm}
	\caption{Visual comparison of the enhanced images by each method on the noisy LOL dataset  (PSNR/SSIM values are also reported).}
	\vspace{-0.2cm}
	\label{fig:8}
\end{figure*}
\begin{table*}[t]
	\caption{ \centering The numerical results in terms of PSNR, SSIM, CSE ($10^3$) and MAE ($\%$) of all competing methods on the noisy LOL datasets with different noise levels, where the  \textbf{bold} denotes the best. }
	\vspace{-0.3cm}
	\resizebox{\textwidth}{26mm}{
		\begin{tabular}{ccccccccccccc}
			\hline
			\multirow{2}*{Noise Levels}  &\multirow{2}*{Eval} & \multicolumn{11}{c}{Methods} \\  \cline{3-13} &~&RetinexNet&R2RNet&KinD++&ZeroDCE++&RUAS&SCI&EnlightenGAN&DCC-Net&SNR&LLFlow&\textbf{Ours}\\
			\hline\hline
			\multirow{4}{*}{$ \sigma=0$} &PSNR$\uparrow$&16.82&20.20&20.92&16.11&11.32&14.86&18.32&22.15&22.96&\textbf{24.83}&24.38\\
			&SSIM$\uparrow$&0.43&0.82&0.80&0.53&0.48&0.61&0.64&0.82&0.86&\textbf{0.89}&\textbf{0.89}\\
			&CSE$\downarrow$&6.36&11.66&5.24&24.68&34.43&33.79&10.10&3.78&4.75&\textbf{2.07}&2.14\\
			&MAE$\downarrow$&14.93&11.68&8.83&11.59&28.23&19.18&13.71&8.96&7.64&\textbf{6.59}&7.13\\
			\hline\hline
			\multirow{4}{*}{$ \sigma=10$} &PSNR$\uparrow$&16.67&18.01&15.16&10.18&10.68&11.98&14.89&15.35&21.72&22.28&\textbf{22.62}\\
			&SSIM$\uparrow$&0.35&0.66&0.62&0.10&0.12&0.10&0.23&0.60&0.77&0.78&\textbf{0.79}\\
			&CSE$\downarrow$&27.98&32.87&7.12&42.60&44.81&46.13&11.09&17.35&3.41&4.32&\textbf{2.50}\\
			&MAE$\downarrow$&12.94&11.58&17.26&26.43&27.85&22.38&15.88&15.18&8.24&7.96&\textbf{7.25}\\
			
			\hline\hline
			\multirow{4}{*}{$ \sigma=20$} &PSNR$\uparrow$&15.04&16.88&16.64&8.30&9.91&10.05&13.21&14.11&20.15&19.29&\textbf{20.99}\\
			&SSIM$\uparrow$&0.18&0.64&0.66&0.04&0.05&0.05&0.12&0.49&0.72&0.70&\textbf{0.74}\\
			&CSE$\downarrow$&48.27&138.34&8.62&62.58&50.68&52.80&13.89&20.29&2.66&12.41&\textbf{1.07}\\
			&MAE$\downarrow$&15.06&12.94&14.92&32.63&28.27&26.47&16.36&16.63&8.46&8.83&\textbf{8.15}\\
			\hline
	\end{tabular}}
	\label{tab2}
\end{table*}

\section{Experimental Results and Analysis}
To validate the effectiveness of our proposed method, we compare it with previous deep models for the noisy low-light image enhancement.
In our experiments, ten closely-related popular deep methods are included for comparison, i.e., RetinexNet~\cite{Wei2018DeepRD}, R2RNet~\cite{Hai2021R2RNetLI}, KinD++~\cite{Zhang2021BeyondBL}, ZeroDCE++ ~\cite{Li2022LearningTE}, EnlightenGAN~\cite{Jiang2021EnlightenGANDL}, RUAS~\cite{Liu2021RetinexinspiredUW}, SCI~\cite{Ma2022TowardFF}, SNR~\cite{Xu2022SNRAwareLI}, DCC-Net~\cite{Zhang2022DeepCC} and LLFlow~\cite{Wang2022LowLightIE}. 
Three benchmark databases, i.e., LOL~\cite{Wei2018DeepRD}, LSRW~\cite{Hai2021R2RNetLI} and RELLISUR datasets~\cite{Aakerberg2021RELLISURAR}, and some real-world images taken by mobile devices at night are used for the evaluations.

For implementing our proposed RLED-Net, the number of the CST layers in each encoder-decoder pair and FRB is 4.
We set the learning rate to 0.0001 and the trade-off parameter empirically to $ \lambda = 0.1 $. 
The rank \emph{r} in LSRB is set to 8 by ablation study. 
For the performance evaluations, four widely-used metrics, i.e., peak signal-to-noise ratio (PSNR), structural similarity (SSIM), mean absolute error (MAE) and Color-Sensitive Error (CSE)~\cite{Zhao2022CRNetUC} are chosen. Different from the PSNR, SSIM and MAE metrics, CSE is a chromatic aberration quantitative metric that directly measures the color difference between image pairs. Similar to MAE, the smaller the CSE value is, the better the enhancement result it indicates. All experiments are carried out and compared on a PC with two 2080Ti GPUs. 

\subsection{ Evaluation on RELLISUR Dataset}
We first evaluate each method on RELLISUR dataset~\cite{Aakerberg2021RELLISURAR} that consists of both indoor and outdoor scenes captured by a DSLR camera. It has 850 low-light/normal-light image pairs and each low-light image corresponds to five under-exposed images. In this study, we adopt the low-light images with exposure value being -3.5 for training and testing. All images are resized into 640×320 for experiments. To verify the enhancement and denoising ability of the evaluated models, Gaussian noise with noise level $ \sigma = 10$ is added into the low-light images. 

The numerical results are described in Table~\ref{tab1}, and some visualization results of enhanced images are shown in Figure~\ref{fig:7}.
From the experimental results, we make two observations.
1) It can be seen from Table~\ref{tab1} that our RLED-Net achieves higher PSNR and SSIM values, and smaller MAE and CSE values than others, followed by DCC-Net, SNR and LLFlow that obtain better results than other remaining methods. That is, RLED-Net can handle the noisy low-light images better and is more robust to noise than competitors.
2) From the visual results, the enhanced images of other methods still have some speckle noise and blur; in contrast, the restored image of our RLED-Net contains less speckle noise and blur. 
We also see that the color of enhanced image by our method is the closest to that of the ground-truth image, and the CSE is the lowest, i.e., RLED-Net produces smaller chromatic aberration. 

\subsection{Evaluation on LSRW Dataset}
We then evaluate each model based on LSRW dataset~\cite{Hai2021R2RNetLI} that contains 5,650 paired images captured by a Nikon D7500 camera and a HUAWEI P40 Pro mobile phone. We take 2,480 paired images taken by Huawei mobile phone for training and 30 paired images for testing. We resize all the images into 960×640 for the shift-windows transformer. To simulate the real dark night and evaluate the robustness of each method against noise, Gaussian noise with $\sigma=10$ is included into the low-light images. The numerical evaluation results are shown in Table~\ref{tab1}, from which similar observations can be obtained as those on  RELLISUR dataset. That is, RLED-Net outperforms other competitors for real-world LLIE in terms of four metrics. The three recent methods, i.e. DCC-Net, SNR and LLFlow,  also obtain highly comparable results on this dataset. 

\subsection{Evaluation on LOL Dataset}
We also evaluate each model based on the LOL dataset~\cite{Wei2018DeepRD} that contains 485 paired low-light/normal-light images for training and 15 paired images for testing. In this study, we resize all the images into 640×320 for each method. To simulate the real dark night and evaluate the robustness of each method against noise, we add different levels of Gaussian noise ($\sigma =0,10,20 $) into the low-light images. 
The numerical enhancement results are described in Table~\ref{tab2} and some visual enhanced results with noise level $\sigma =10$ are shown in Figure~\ref{fig:8} as examples. We have the following findings. 1) From the numerical results in Table~\ref{tab2}, we find when the original LOL dataset is used, i.e., noise level$\sigma =0$, LLFlow obtains the best result, followed closely by our RLED-Net. However, when the noise levels increase, the performance of LLFlow drops faster than our RLED-Net, and our method can outperform LLFlow in the noisy cases. 2) From the visualizations in Figure~\ref{fig:8}, our RLED-Net obtains better enhancement results, with less noise and more accurate detail-recovery effect. In contrast, other compared methods produce blurry recoveries due to interference of noise in low-light images, and the recovered images still contain much more noise. 

\begin{figure*}[t]
	\centering
	\includegraphics[width=\linewidth]{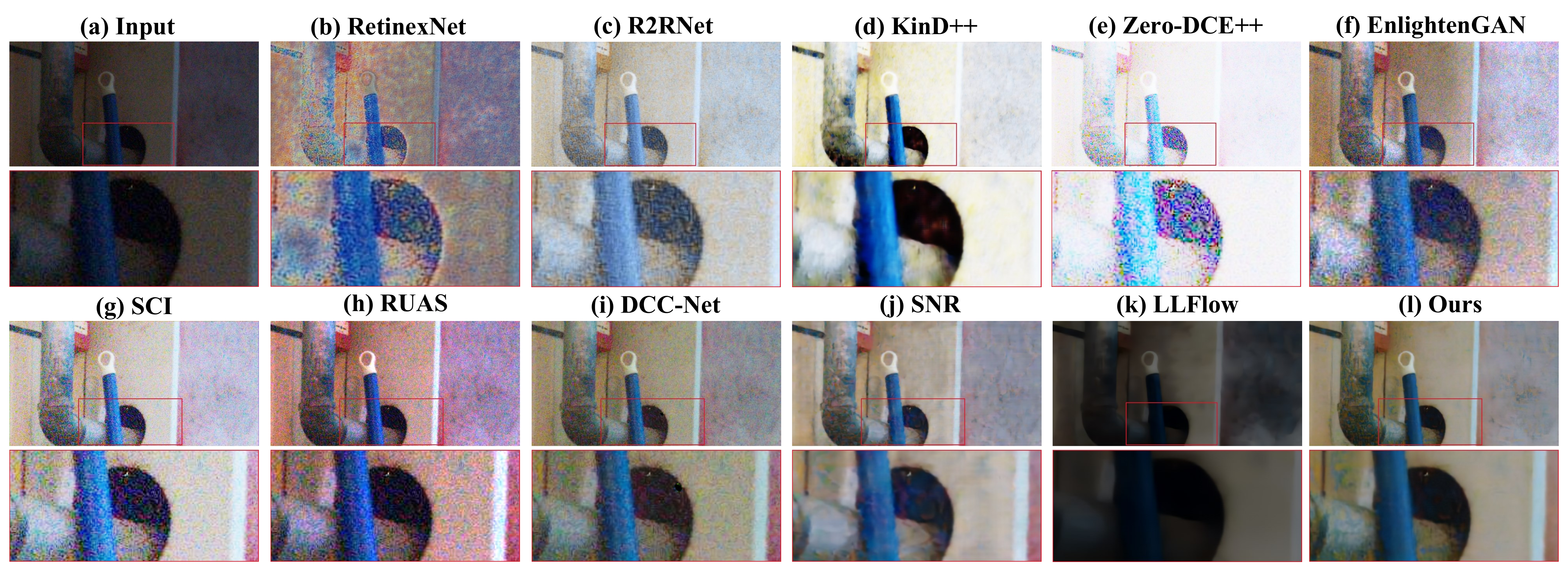}
	\vspace{-0.6cm}
	\caption{LLIE comparison of each method based on a real-world noisy low-light image (with real noise).}
	\vspace{-0.4cm}
	\label{fig:9}
\end{figure*}

\begin{figure}[t]
	\centering
	\includegraphics[width=\linewidth]{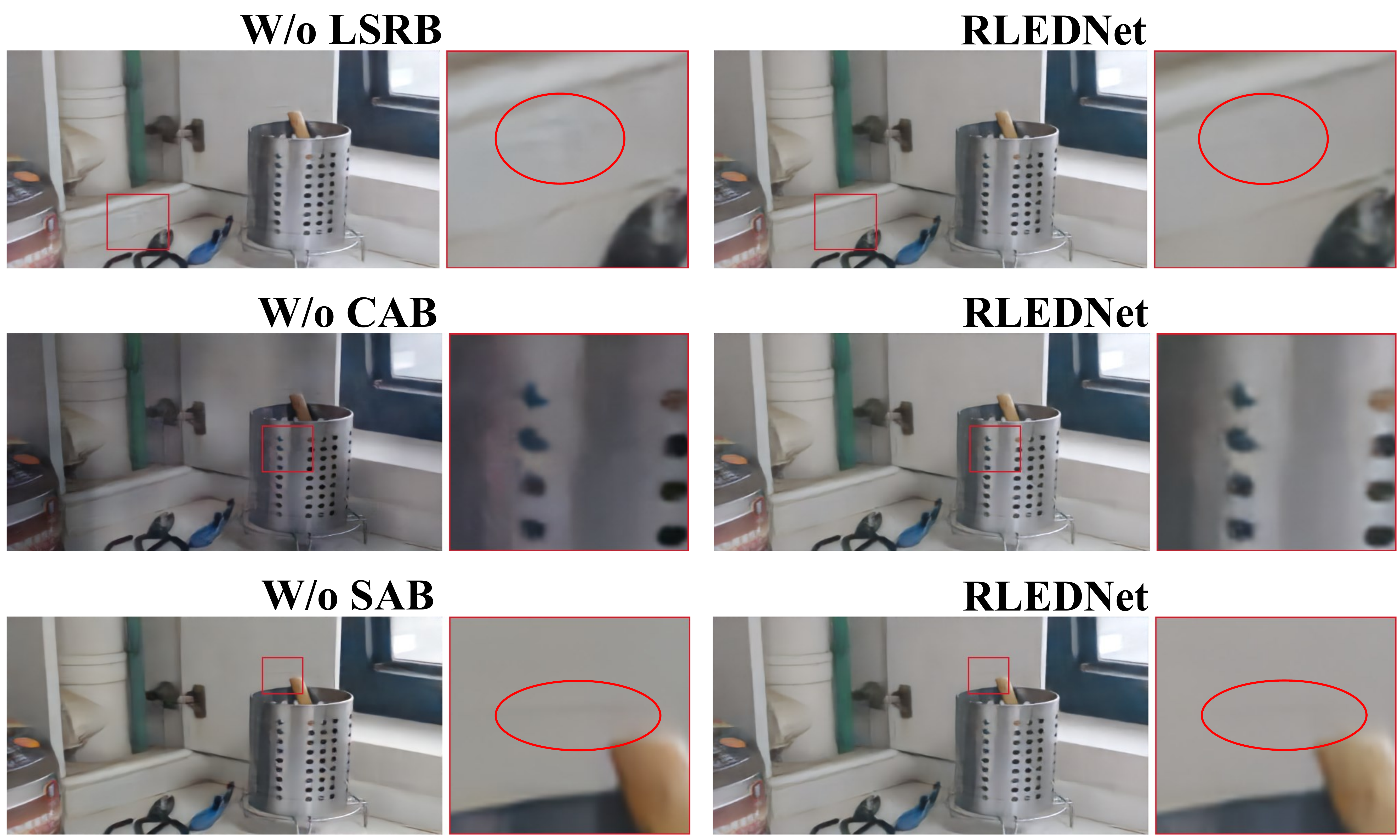}
	\vspace{-0.7cm}
	\caption{Visual comparison of different components in our model.}
	\vspace{-0.1cm}
	\label{fig:10}
\end{figure}
\subsection{Evaluation on Real-world Low-Light Images}
To evaluate the generalization ability of each deep LLIE model, we also use some real-world low-light images for evaluations. Due to the lack of paired images for training, each deep model is trained on the LOL dataset with Gaussian noise level $\sigma=10$, and then tested on real low-light images. The visual results are shown in Figures~\ref{fig:1} and \ref{fig:9}. It can be seen that our RLED-Net handles the real-world LLIE task better by effectively removing the noise in real images. More importantly, our method can retain more detailed texture information, which means our model has a stronger generalization ability to real cases. In contrast, the enhanced images by other methods have much more noise, and are still blurred due to the influence of noise. 

\begin{table}[t] 
	\centering
	\caption{\centering Comparison on the LOL dataset with varying ranks.}
	\vspace{-0.3cm}
	\resizebox{\columnwidth}{8.5mm}{
		\begin{tabular}{ccccc}
			\hline
			Model&W/o LSRB&W/o SAB&W/o CAB&RLED-Net\\
			\hline
			PSNR$\uparrow$&22.15&22.41&16.91&\textbf{22.62}\\
			SSIM$\uparrow$&0.77&0.78&0.72&\textbf{0.79}\\
			MAE($\%$)$\downarrow$ &7.98&7.35&15.51&\textbf{7.25}\\
			\hline
		\end{tabular}
	}
	\label{tab3}
	\vspace{-0.2cm}
\end{table}

\begin{table}[t] \scriptsize
	\centering
	\caption{\centering Comparison on the LOL dataset with varying ranks.}
	\vspace{-0.3cm}
	\resizebox{\columnwidth}{8.5mm}{
		\begin{tabular}{cccccc}
			\hline
			Rank&2&4&8&16&32\\
			\hline
			PSNR$\uparrow$&21.25&21.37&\textbf{22.62}&21.87&20.42\\
			SSIM$\uparrow$&0.78&0.78&\textbf{0.79}&0.79&0.78\\
			MAE($\%$)$\downarrow$&7.81&\textbf{8.09}&7.25&7.71&9.03\\
			\hline
		\end{tabular}
	}
	\label{tab4}
	\vspace{-0.3cm}
\end{table}

\subsection{Ablation Studies}
We evaluate the impact of each component in our RLED-Net on the final performance, i.e., LSRB, SAB and CAB. 
To demonstrate the effectiveness of the structure design of LSRB, we remove the latent subspace decomposition process and reserve the shallow feature extraction. To prove the effectiveness of the components SAB and CAB in CST, we use them separately to construct the encoder-decoders and keep the number of channels unchanged. The ablation analysis results are reported in Table~\ref{tab3}, where \textbf{W/o LSRB} denotes RLED-Net without LSRB, \textbf{W/o SAB} and \textbf{W/o CAB} denote the encoder-decoders without SAB and CAB, respectively. In addition, we also exhibit some visual results in Figure~\ref{fig:10} to see the effectiveness of each component more straightforwardly. In this study, LOL dataset with noisy level $\sigma=10$ is used. We conclude that when LSRB is removed, some areas of the restored image are blurred, due to the noise; when CAB is removed, the brightness of the restored image becomes uneven, which may be caused by the loss of global information; when SAB is removed, some local parts become inaccurate. These results prove that each component is important for ensuring the performance of our method. 
In addition, the designed LSRB module also involves the selection of the rank of matrices, so we also evaluate it by varying the ranks of the decomposed matrix, and report the experimental results in Table~\ref{tab4}. We find that RLED-Net performs the best when the rank is 8. 

\section{Concluding Remarks}
We explored the challenging task of seeing through the noisy dark in the sRGB color space, and proposed a real-world low-light enhancement \& denoising network RLED-Net. To enable our model to generalize well to real-world scenarios commonly with various noise, we designed a plug-and-play and differentiable latent subspace reconstruction block by characterizing the low-rank structures of images, so that redundant information and noise can be removed. We also designed a novel CST layer with two parallel branches and based on it, a CSTNet and a simple feature refine block have been built to simultaneously reduce the loss of global color/shape information and local edge/texture information caused by the noise and low-resolution pixels, and finally refine the features. 

Extensive experiments on benchmark datasets have demonstrated the effectiveness of our RLED-Net for low-light enhancement and denoising, and the strong generalization ability to real low-light scenes. In the future, we will explore more complex real-world LLIE tasks, for instance, LLIE with extreme non-uniform light or rainy night.
\vspace{-0.3cm}

{\small
\bibliographystyle{ieee_fullname}
\bibliography{egbib}
}

\end{document}